\documentclass[preprint,12pt]{elsarticle}

\usepackage{amssymb}
\usepackage{amsthm}
\usepackage{amsmath}
\usepackage{algorithm}
\usepackage[noend]{algorithmic}
\usepackage{color}
\renewcommand{\algorithmicprint}[1]{\textbf{break}} 

\newtheorem{theorem}{Theorem}
\newtheorem{definition}{Definition}
\newtheorem{lemma}[theorem]{Lemma}
\newtheorem{proposition}[theorem]{Proposition}
\newtheorem{example}{Example}
\newtheorem{corollary}{Corollary}

\newcommand{\bits}[1]{[#1]}
\newcommand{\ones}[1]{[#1]}
\newcommand{\set}[2]{\left\{#1\vphantom{#2} \, \right| \left.\vphantom{#1}#2\right\}}

\def\H{\textit{ClH}}
\def\B{\{0,1\}^n}
\def\implies{\rightarrow}
\def\entails{\models}

\journal{a special issue of a professional journal} 

\begin{document}

\begin{frontmatter}

\title{Learning Definite Horn Formulas from Closure Queries}

\author[UPC]{Marta Arias\corref{cor1}\fnref{label,label2}}
\ead{marias@cs.upc.edu}

\author[UPC]{Jos\'e L.~Balc\'azar\fnref{label,label2}}
\ead{jose.luis.balcazar@upc.edu}

\author[Unican]{Cristina T\^\i{}rn\u{a}uc\u{a}\fnref{label,label3}}
\ead{cristina.tirnauca@unican.es}

\address[UPC]{LARCA Research Group, Department of Computer Science,\\ Universitat Polit\`ecnica de Catalunya, Barcelona, Spain}
\address[Unican]{Departamento de Matem\'aticas, Estad\'istica y Computaci\'on,\\ Universidad de Cantabria, Santander, Spain}

\cortext[cor1]{Corresponding author}
\fntext[label]{Partially supported by project BASMATI (TIN2011-27479-C04-04) of Programa Nacional de Investigaci\'on, Ministerio de Ciencia e Innovaci\'on (MICINN), Spain.}
\fntext[label3]{Partially supported by project PAC::LFO (MTM2014-55262-P) of Programa Estatal de Fomento de la Investigaci\'on Cient\'ifica y T\'ecnica de Excelencia,
Ministerio de Ciencia e Innovaci\'on (MICINN), Spain.}

\fntext[label2]{Partially supported by grant 2014SGR 890 (MACDA) from AGAUR, Generalitat de Catalunya.}

\begin{abstract}
A definite Horn theory is a set of $n$-dimensional Boolean vectors whose characteristic function is   
expressible as a definite Horn formula, that is, as conjunction 
of definite Horn clauses. The class of definite Horn theories is known to be learnable under different 
query learning settings, such as learning 
from membership and equivalence queries 
or learning from entailment. 
We propose yet a different type of query: the closure query. 
Closure queries are a natural extension of membership queries 
and also a variant, appropriate in the context of definite Horn formulas, 
of the so-called correction queries. 
We present an algorithm that learns conjunctions of definite Horn clauses 
in polynomial time, using  closure and equivalence queries, and 
show how it relates to the canonical Guigues-Duquenne basis
for implicational systems. We also show how the different query 
models mentioned relate to each other by either showing full-fledged 
reductions by means of query simulation (where possible), or by 
showing their connections in the context of particular algorithms 
that use them for learning definite Horn formulas.

\end{abstract}

\begin{keyword}
Query learning \sep definite Horn clauses \sep closure operators

\end{keyword}

\end{frontmatter}

\section{Introduction}

Query Learning \cite{AngluinQaCL} is one of the most well-known and studied
theoretical learning models available. According to this model, a learning agent
or algorithm interacts with the world by asking questions (that is, \textit{queries}) to one or several oracles that reveal partial information about an unknown but previously fixed concept (the \textit{target}). Learning is achieved when the learning agent outputs a representation of the unknown concept. Naturally, as it is customary in a computational context, the number of queries that the algorithms need to make in order to learn is a resource that one needs to account for and restrict; efficient algorithms in the Query Learning context should address to the oracle a polynomial amount of queries. 
Finally, a concept class is said to be \textit{learnable} within a query learning model if there exists an algorithm that, by asking a polynomial number of its available queries,  is able to discover any previously fixed concept from the concept class.

Query Learning thus deals with determining learnability of different concept classes under different variants of learning models (that is, with different query types). The more complex a concept class is, the more queries will in general be necessary to learn it. And one concept class may be learnable under a particular set of query types but not others (see e.g. \cite{angluin1995won} for examples of this).
Several extensions, of different focus and generality, appeared
subsequently. One very general notion of query learning
is that of \cite{BCGKL}; in a less general level, some of
these different extensions are of interest for this paper.

In this paper, we focus our attention to the class of (propositional) definite Horn formulas. The seminal paper by Angluin, Frazier and Pitt \citep{AngluinFrazierPitt} established that this class is indeed query learnable\footnote{To be more precise, their algorithm learns the more general class of Horn formulas.}.
The types of queries that they allowed are the most commonly used in query learning \cite{AngluinQaCL}: standard membership queries (SMQs) and standard equivalence queries (SEQs). In an SMQ, the algorithm asks whether a particular example (a truth assignment in our case) belongs to the unknown target concept, and the answer it receives is YES if the assignment is satisfied and NO otherwise. In an SEQ, the algorithm asks whether a particular Horn formula (the hypothesis) is semantically equivalent to the target formula, and the answer it receives is YES if it is indeed so and NO otherwise. In the case of a negative answer, a \textit{counterexample} is provided, that is, an assignment that is satisfied by the target but not by the hypothesis or vice versa. The counterexample is then used by the algorithm to refine its hypothesis and learning continues.


Since the introduction of the first query learning models \cite{AngluinQaCL}, several further variants of query types have been introduced and studied.
One is the family of Unspecified Attribute Value queries
\cite{GoldmanKwekScott},
the query-learning parallel to the Restricted Focus
of Attention extension to the PAC model \cite{BDD}.
Its key traits are the context of $n$-dimensional Boolean
vectors (for fixed $n$) and the ability to handle each
dimension somehow ``individually'', by means of the use
of ``don't-care'' symbols to allow the query to focus
on specific dimensions. 

A second variant, also working on
$n$-dimensional Boolean vectors but moving to a slightly
more abstract notion of query, is proposed in
\cite{FrazierPitt}: the entailment query, 
where simple formulas (in that concrete case, Horn
clauses) play the role of individual examples.
In this protocol, in an entailment membership query (EMQ) 
the learner proposes a 
Horn clause
and receives, as answer, a Boolean value indicating 
whether it is entailed by the target, seen as a 
propositional 
(in that case, Horn)
formula. 
Similarly, in the entailment equivalence query (EEQ), a Horn formula
is proposed and, in case of a negative answer, the
provided counterexample is a Horn clause that is
entailed by exactly one of the two formulas, the
target and the query.

Yet a different variant of entailment query is employed in \cite{GanterAE}
for the algorithm known as Attribute Exploration, also in a
context very similar to learning Horn clauses. This is a
protocol where the query is an implication, that is, a 
conjunction of clauses sharing the same antecedent; the 
main difference is as follows: one gets
as answer either YES if the implication is entailed by the
target, or a counterexample that satisfies the target but 
not the implication. Thus, this variant is midway through 
between plain entailment membership, to which it resembles
most, and standard equivalence, because a counterexample
assignment is received in the negative case.

Finally, in \cite{BeDeTi06,Tirn08,Tirn09,TirKob09},  
we find a different extension: Correction Queries,
which model a very intuitive idea from linguistics: 
instead of a simple NO answer as in the case of 
SMQs, the teacher provides a ``correction'', 
that is, an element of the target language at minimum distance 
from the queried example.  

One must note that, whereas several positive results prove
that the availability of certain query combinations allows for
polynomial-time learnability, there are negative results that
show that many representation classes are impossible to learn
from polynomially many individual queries like membership or 
equivalence \cite{AngluinQaCL,angluin1995won}.
 
This paper's contributions are two-fold.
First, we propose 
a quite natural notion of \emph{closure query} (CQ)
and a polynomial time
learning algorithm for definite Horn Boolean functions via equivalence
and closure queries;
it is closely related to the algorithms 
in \cite{AngluinFrazierPitt} and \cite{FrazierPitt}.
The second part of the paper studies the relationships 
between our newly introduced closure query model and 
other well-known query models. While some of the 
relationships are already known (and these are duly noted), 
we obtain interesting new ones. More precisely, our
algorithm yields back, through a quite intuitive 
transformation, the algorithm for Learning from Entailment \cite{FrazierPitt}. 
Additionally, as we shall see as well, also the
celebrated algorithm to learn definite Horn theories from
membership and equivalence queries of \cite{AngluinFrazierPitt}
can be related to this approach, in that the usage it makes
of positive examples can be understood as progressing towards
the identification of unavailable closures. We believe that 
these connections we develop provide insight, and also help 
in establishing an overview of the 
strengths and weaknesses of all the variants under study.
In addition to the relation of 
the learning algorithm proposed in the first half of the 
paper to these already existing variants, 
further relationships between models are shown to be 
possible in the more general form of query simulation.

Closure queries share a number of traits with each of the
query models discussed before.
Under a natural notion of ``correcting upwards'' (see below
for precise definitions),
in the context of a definite Horn target, the closest correction to
a negative example is exactly its \emph{closure} under the target.
Thus, we find a variant of correction query for definite Horn targets
that allows for (limited) manipulation of individual dimensions of
the Boolean hypercube, as do the other query models we have mentioned,
and provides an explanation of ``what are we looking for''
along both the processes of entailment queries and the positive
examples of the learning algorithm from membership and equivalence
queries. Our advances are based on
the novel view on Horn learning via queries deployed
more recently in \cite{AriasBalcazarALT,AriasBalcazarML}.
The introduction of the closure query is, in fact, motivated by these
two papers in the following
sense: the main aim in revisiting the original 
algorithm \cite{AngluinFrazierPitt} was to improve on its query complexity. 
This objective is still unfulfilled; however, we believe that the closure query
provides the ``right amount'' of expressiveness and information to the algorithm in order to capture the essence of difficulty in learning, while at the same time making some of the book-keeping details easier to deal with. Thus, working with the closure query model, we believe that we are in a better position to  
answer the fundamental question of whether the original algorithms of \cite{AngluinFrazierPitt,FrazierPitt} are indeed optimal or can otherwise be improved upon.

\section{Preliminaries}

We work within the standard framework in 
propositional 
logic, where one is given
an indexable set of propositional variables of cardinality $n$, 
Boolean functions are subsets of the Boolean hypercube $\B$, 
and these functions are represented by logical formulas over 
the variable set in the standard way. Binary strings of length~$n$
assign a Boolean value for each variable, and are therefore
called assignments; given any Boolean function or formula $H$,
the fact that assignment $x$ makes it true (or ``satisfies'' it)
is denoted $x\models H$.
Following the standard overloading of the operator, $H\models H'$
means that, for every assignment $x$, if $x\models H$ then $x\models H'$.
Assignments are partially ordered bitwise according to $0\leq1$
(the usual partial order of the hypercube);
the notation is $x\leq y$.
%

%

A literal is a variable or its negation.
A conjunction of literals is a \emph{term}, 
and if none of the literals appears negated it is a \emph{positive term},
also often referred to as a \emph{monotone term} or \emph{monotone conjunction}. 
We often identify positive terms and mere sets 
of variables; in fact, we switch back and forth between set-based 
notation and assignments. We denote variables with letters from the beginning of the alphabet ($a,b,c, ..$), terms, 
or equivalently subsets of variables, 
with Greek letters ($\alpha, \beta, ..$) and assignments with letters from the end of
the alphabet ($x,y,z,..$). We may abuse notation at times and it should be 
understood that if we use a subset $\alpha$ when an assignment 
is expected, it is to be interpreted as the assignment that 
sets to 1 exactly those variables in $\alpha$. We denote 
this explicitly when necessary by $x = \bits{\alpha}$.
Similarly, if we use an assignment $x$ where a subset of variables 
is expected, it is to be understood that we mean the set of variables 
that are set to 1 in $x$. We denote this explicitly 
by $\alpha = \ones{x}$.
Clearly, we have a bijection between sets of propositional 
variables and assignments, and $x = \bits{\ones{x}}$ and 
$\alpha = \ones{\bits{\alpha}}$ for all assignments $x$ 
and variable sets $\alpha$.

\subsection{Horn Logic}

In this paper we are only
concerned with definite Horn functions, and their representations using
conjunctive normal form (CNF).
A Horn CNF formula is a conjunction of Horn clauses. A clause is a disjunction
of literals. A clause is \emph{definite 
Horn} if it contains exactly one positive literal, and it is \emph{negative}
if all its literals are negative. A clause is \emph{Horn} if it is either definite Horn
or negative. 

Horn clauses are generally viewed as implications where the 
negative literals form the antecedent of the implication 
(a positive term), and the singleton consisting of the positive
literal, if it exists, forms the consequent of the clause. 
As just indicated, along this paper it will always exist.

An implication $\alpha\implies\beta$,
where both $\alpha$ and $\beta$ are sets of propositional variables
with $\alpha$ possibly empty, but not $\beta$,
is to be interpreted as the conjunction of definite Horn clauses 
$\bigwedge_{b\in\beta}\alpha\rightarrow b$.
A semantically equivalent interpretation is to see both sets of
variables $\alpha$ and $\beta$ as positive terms; the Horn formula in
its standard form is obtained by distributivity over the variables of $\beta$.
Of course, any result that holds for Horn formulas in 
implicational form with no other restrictions also holds 
for the clausal representation unless it explicitly depends
of the implications proper, such as counting the number of
implications, as we will do below.
Furthermore, we often use sets to denote conjunctions, as we do with 
positive terms, also at other levels: 
a generic (implicational) CNF $\bigwedge_i (\alpha_i\implies\beta_i)$ 
is often denoted in this text by $\{(\alpha_i\implies\beta_i)\}_i$.
Parentheses are mostly optional and generally used
for ease of reading.

An assignment $x\in\B$ satisfies the implication 
$\alpha\implies\beta$, denoted
$x\models\alpha\implies\beta$, if it either falsifies the antecedent 
or satisfies the consequent, that is, $x\not\models\alpha$ or
$x\models\beta$ respectively, where now we are interpreting both
$\alpha$ and $\beta$ as positive terms ($x\models\alpha$ if and only if $\alpha\subseteq\ones{x}$ if and only if $\ones{\alpha}\leq x$, see Lemma 1 of \cite{AriasBalcazarML}). 

Not all Boolean functions are Horn. 
The following semantic characterization is a well-known
classic result of \cite{Horn1956,McKinsey1943},
proved in the context of propositional Horn logic 
e.g. in \cite{KhardonRo1996}:

\begin{theorem}[\cite{Horn1956,McKinsey1943,KhardonRo1996}]
A Boolean function admits a Horn CNF representation if and only if
the set of assignments that satisfy it is closed under 
bit-wise intersection.
\end{theorem}

A Horn function admits several syntactically 
different Horn CNF representations; 
in this case, we say that these
representations are equivalent. 
Such representations are also known as \emph{bases} for the Boolean function they represent. 
The \emph{size} of a Horn function is the minimum number
of clauses that a Horn CNF representing it must have. 
The \emph{implication size} of a Horn function is defined 
analogously, but allowing formulas to have implications 
instead of clauses. Clearly, every definite clause can be phrased as an implication, 
and thus the implication size of a given Horn function
is always at most that of its standard size as measured 
in the number of clauses.

Horn CNF representations may as well include unnecessary
implications. We will need to take this into account: an implication
or clause
in a Horn CNF $H$ is {\em redundant} if it can be removed 
from $H$ without changing the Horn function represented. 
A Horn CNF is {\em irredundant} or {\em irreducible} if it does not contain any
redundant implication or clause. Notice that an irredundant $H$ may
still contain other sorts of redundancies, such as implications with consequents
larger than strictly necessary. 


\subsection{Closure Operator and Equivalence Classes}
\label{ss:closop}

We will employ the well-known method of 
\emph{forward chaining} for definite Horn 
functions; see e.g.~\cite{buning1999propositional}.
Given a definite Horn CNF $H=\{\alpha_i\implies\beta_i\}_i$ 
and an initial subset of propositional variables $\alpha$, 
we can construct a chain of subsets of propositional 
variables by successively adding right-hand sides of implications,
provided that the corresponding left-hand side is 
already contained in the current subset. Given
a set of variables $\alpha$, the maximal outcome
of this process is denoted $\alpha^{\star}$,
and contains all the variables ``implied''
by the set of variables $\alpha$. As it is well-known,
$\alpha^{\star}$ is well-defined, and only depends 
on the Boolean function represented by $H$, not on 
the representation $H$ itself.
The corresponding process on assignments provides
the analogous operator $x^{\star}$.

Note that the closure operator is defined with respect
to a function which is not explicitly included in the notation $x^{\star}$.
It should be clear from the text, however, with respect to what function the closure is taken.


It is easy to see that the $\star$ operator is extensive 
(that is, $x\leq x^{\star}$ and $\alpha\subseteq \alpha^{\star}$),
monotonic 
(if $x\leq y$ then $x^{\star}\leq y^{\star}$, 
and if $\alpha\subseteq	\beta$ then $\alpha^{\star}\subseteq \beta^{\star}$) 
and idempotent 
($x^{\star\star}=x^{\star}$, and $\alpha^{\star\star} = \alpha^{\star}$) 
for all assignments $x,y$ and variable sets $\alpha,\beta$; 
that is, $\star$ is a closure operator. 
Thus, we refer to $x^{\star}$ as the \emph{closure} of $x$ 
w.r.t. a definite Horn function $f$.
An assignment $x$ is said to be \emph{closed} iff $x^{\star}=x$, 
and similarly for variable sets. The following holds for
every definite Horn function $f$ (see Theorem 3 in \cite{AriasBalcazarML}):

\begin{proposition}[\cite{AriasBalcazarML}]
\label{p:closeds}
Let $f$ be a definite Horn function; let $\alpha$ be an arbitrary variable subset, $b$ any variable and $x$ an arbitrary assignment. Then,
\begin{enumerate}
\item $f \entails \alpha \implies b$ if and only if $b \in  \alpha^{\star}$,
\item $x=x^{\star}$  if and only if $x \entails f$,
\item $x^{\star}=\wedge\{y \mid x \leq y \textrm{ and } y \entails f\}$,
\end{enumerate}
\end{proposition}

Therefore, $f \entails \ones{y} \implies \ones{y^{\star}}$ whenever the closure of $y$ is computed with respect to $f$. Moreover, for any assignment $x$, there is a uniquely defined assignment $y$ evaluated positively by $f$ with $y$ bitwise minimal such that $x\leq y$, namely, $y = x^{\star}$.

For a fixed function $f$, this closure operator induces a partition over the set 
of assignments $\B$ in the following straightforward way: 
two assignments $x$ and $y$ belong
to the same class if $x^{\star} = y^{\star}$, where both closures are taken w.r.t. $f$. 
This notion of equivalence class carries over as expected to the power set
of propositional variables: the subsets $\alpha$ and $\beta$ belong
to the same class if $\alpha^{\star} = \beta^{\star}$. 
It is worth noting that each equivalence class consists of 
a possibly empty set of assignments that are not closed and 
a single closed set, its representative.

Furthermore, the notion of equivalence classes carries over to implications
by identifying an implication with its antecedent. Thus, two 
implications belong
to the same class if their antecedents have the same closure (w.r.t. a fixed $f$). Thus, the class
of an implication $\alpha\implies\beta$ is, essentially, $\alpha^{\star}$.

\begin{example}
This example is taken from \cite{GD}.
Let $H = \{e \implies d,bc \implies d, bd \implies c, cd \implies b, ad \implies bce, ce \implies ab\}$.
Thus, the propositional variables are
$a,b,c,d,e,f$.
The following table illustrates the partition induced by the equivalence classes on the 
implications of $H$, where closures are taken with respect to $H$ itself.
The first column is the 
implication identifier, the second column is the 
implication itself, and the third column
corresponds to the class of the 
implication.
As one can see, there are three equivalence classes: one containing the first 
implication, another one
containing 
implications 
2, 3, and 4; and a final one containing 
implications 5 and 6.

\begin{center}
\begin{tabular}[h]{l l l}
\hline
1\ \ \ & \ $e \implies d$    & \ \ $ed$    \\
\hline
2\ \ \ & \ $bc \implies d$   & \ \ $bcd$   \\
3\ \ \ & \ $bd \implies c$   & \ \ $bcd$   \\
4\ \ \ & \ $cd \implies b$   & \ \ $bcd$   \\
\hline
5\ \ \ & \ $ad \implies bce$ & \ \ $abcde$ \\
6\ \ \ & \ $ce \implies ab$  & \ \ $abcde$ \\
\hline
\end{tabular}
\end{center}
\end{example}

\subsection{A Related Closure Operator}

Now we proceed to define another important operator which is similar
in flavor to the closure operator $\star$ seen above. 

Let $H$ be any definite Horn CNF, and $\alpha$ any variable subset.
Let $H(\alpha)$ be those implications of $H$ whose antecedents 
fall in the same equivalence class as $\alpha$, namely,
$H(\alpha) = \set{\alpha_i\implies\beta_i}{\alpha_i\implies\beta_i \in H \hbox{ and } \alpha^{\star} = \alpha_i^{\star}}.$ The closure is taken
with respect to $H$ itself.

Given a definite Horn CNF $H$ and a variable subset $\alpha$, 
we introduce a new operator $\bullet$ \cite{GD,Maier1980,Wild}
that we define as follows:
$\alpha^{\bullet}$ is the closure of $\alpha$
\emph{with respect to the subset of implications $H \setminus H(\alpha)$}.
That is, in order to compute $\alpha^{\bullet}$ one does 
forward chaining starting with $\alpha$ \emph{but one is not
allowed to use implications in $H(\alpha)$}.


\begin{example}
Let $H = \{a\implies b, a\implies c, c\implies d\}$. 
Then, $(ac)^{\star} = abcd$ but $(ac)^{\bullet} = acd$ 
since $H(ac) = \{a\implies b, a\implies c\}$ and we 
are only allowed to use the implication $c\implies d$ when
computing $(ac)^{\bullet}$.
\end{example}

This new operator is, in fact, a closure operator,
well-known in the field of Formal Concept Analysis; there,
assignments that are closed with respect to it are sometimes
called \emph{quasi-closed}.

\subsection{Saturation and the Guigues-Duquenne Basis}

In this section we review briefly part of our results from our previous work \cite{AriasBalcazarML}.
We will skip many details as they can be found in the aforementioned article.
These results are, in fact, an interpretation of the work of \cite{GD,Wild} which were stated
in the context of formal concepts, closure systems and lattices.

We say that an implication $\alpha\implies\beta$ of a definite Horn CNF $H$ is 
\begin{itemize}
\item \emph{left-saturated} if $\alpha = \alpha^{\bullet}$ (the quasi-closure is taken with respect to $H$)
\item \emph{right-saturated} if $\beta = \alpha^{\ast}$ (the closure is taken with respect to $H$)
\item \emph{saturated} if it is both left and right-saturated (the closure is taken with respect to $H$)
\end{itemize}

Then, a definite Horn CNF is saturated if all of its implications are. 
Moreover, any saturated definite Horn CNF must be irredundant (see Lemma 2 of \cite{AriasBalcazarML} for a proof). A result from \cite{GD,Wild} states that

\begin{theorem}[\cite{GD,Wild}]
Definite Horn functions have at most one saturated basis, which is of minimum implicational size. This basis is called the Guigues-Duquenne 
(GD) basis.
\end{theorem}

See \cite{AriasBalcazarML} for additional discussion.
In particular, there we show that the GD basis of a 
given definite Horn CNF representation can be 
computed with the following procedure: 
using forward chaining, right-saturate every clause. Then, use 
forward chaining again to compute the left-saturation of the 
left hand sides of the implications using the information on 
the equivalence classes of the existing implications to do the 
left-saturation properly. Finally, remove all those clauses that 
are redundant. 

\subsection{Closure Queries}

The above leads naturally to a clear notion of closure query (CQ):

\begin{definition}
For a fixed definite Horn formula $T$ on $n$ 
propositional variables and given $n$-bit vector $y$,
the answer to a closure query on input $y$ is~$y^{\star}$, 
that is, the closure of $y$ with respect to $T$.
\end{definition}

This query can be seen as a natural variant of correction query:
under the condition that all corrections are ``upwards'', namely, that they are allowed only to
change a zero into a one, Proposition~\ref{p:closeds} tells us
that, for every assignment $y$ that is negative for $T$, there is a 
unique ``closest'' correction query, and it is exactly $y^{\star}$, the closure of $y$ with respect to $T$.

Closure queries provide us with a way of correctly identifying right hand sides of implications in one shot, since saturated implications are always of the form 
$ \ones{y} \implies \ones{y^{\star}}$. We shall see in the next section that this is exploited during the learning process.

An oracle answering closure queries can be implemented to run in linear time in $|T|$ and $n$ (see Theorem 2 of \cite{AriasBalcazarML}).

\section{Learning Definite Horn Theories from Closure and Equivalence Queries}

As usual in Query Learning, a target definite Horn function $T$ is
fixed, and the learning algorithm interacts with an
environment able to provide information about $T$ in the
form required by the corresponding query protocol. 
In our case, this amounts to the learning algorithm
being able to use at any time the closure $y^{\star}$ of
any Boolean vector $y$ as necessary, as it can be 
obtained from a closure query (the closure is computed with respect to the target $T$). 
Equivalence queries
(denoted as $EQ()$ in the algorithms)
are used in the standard manner as control of termination:
the algorithm finishes exactly when the equivalence
query receives a positive answer, which guarantees
correctness provided that the algorithm is shown to
terminate.

\begin{theorem}
\label{t:hstar}
Definite Horn formulas are learnable from equivalence and
closure queries in polynomial time.
\end{theorem}

\begin{proof}
Each equivalence query will take the form of a hypothesis
definite Horn formula $hyp(N)$ based on a list of $n$-bit vectors $N$; 
namely, it will be a conjunction of implications, defined as follows:
$$
	hyp(N) = \bigwedge_{y\in N} \ones{y}\implies \ones{y^{\star}}
$$
In fact, as we shall see momentarily, $N$ will be a list
of negative examples, as usual in Horn clause learning. 
Clearly, given $N$, $hyp(N)$ can be constructed easily
using closure queries.

We observe now that the combination of an inequality and a
closure leads to a membership query: $z$ is a negative
example if and only if $z < z^{\star}$, because always $z\leq z^{\star}$, and
positive examples are exactly those that coincide with
their closure, by Proposition \ref{p:closeds}.

We combine these ingredients as described in Algorithm~\ref{fig:hstar} which
we will call \H. The proof of its correctness
is built out of the following two lemmas. These lemmas refer to the elements in $N$, the list of counterexamples,
as $y_1, y_2, y_3, \ldots, y_{|N|}$.

\begin{lemma}\label{lemma:negcex}
$T \entails hyp(N)$, therefore counterexamples are 
always negative.
\end{lemma}
\begin{proof}[Proof of Lemma \ref{lemma:negcex}]
Take any assignment $y$. Since the closure $y^{\star}$ is taken with respect
to the theory $T$, we have that $T\entails \ones{y}\implies \ones{y^{\star}}$ for every $y$ and in particular all those
$y\in N$, and therefore,
$T \entails \bigwedge_{y\in N} \ones{y}\implies \ones{y^{\star}} = hyp(N)$ as required.
\end{proof}

\begin{lemma}\label{lemma:z}
For $i<j$, there is a positive $z$ with 
$y_i \land y_j \leq z \leq y_j$, and, therefore,
each $y_i$ violates  different implications of $T$.
\end{lemma}
\begin{proof}[Proof of Lemma \ref{lemma:z}]
%

We argue inductively along the successive updates of $N$.
We need to establish the fact (1) at the time of appending
a new element of $N$,  and (2) we need to argue that refinements to existing $y\in N$ that
take place maintain the fact stated. 
We will show (2) in detail; (1) is proven similarly and so we omit the details.

First note the easiest case whereby $y_i$ gets refined
into $y'_i = y_i\land x$.
This leaves $y_j$ untouched, and brings down $y_i\land y_j$ into 
$y'_i\land y_j$; the same value of ~$z$, given by the induction hypothesis on $y_i, y_j$ before the update, will do:
$y'_i\land y_j \leq y_i\land y_j\leq z \leq y_j$.

Now consider the case in which $y_j$ is refined
into $y'_j = y_j\land x$. We assume as inductive hypothesis
that a corresponding $z$ exists before the refinement:
$y_i\land y_j\leq z \leq y_j$.

We establish first the following auxiliary claim:
there is a positive
example $z'$ for which $y_i \land x \leq z' \leq x$.
To find such $z'$, observe that $y_i$ 
came before $y_j$ but was not chosen for refinement; either
$y_i\land x$ is itself positive, and we can simply 
choose $z' = y_i\land x$,
or $y_i\leq x$. 
Since $x$
was a negative counterexample, it must satisfy the query,
so we have that $x \entails \ones{y_i}\implies \ones{y_i^{\star}}$;
therefore $y_i^{\star}\leq x$ since $y_i\leq x$.
We pick $z' = y_i^{\star}$,
which is of course positive.

At this point, we have 
the already existing $z$ fulfilling
$y_i \land y_j\leq z \leq y_j$,
and the $z'$ just explained for which
$y_i \land x \leq z' \leq x$.
Observe the following:
$y_i \land y'_j = y_i \land y_j \land x =
y_i \land y_i \land y_j \land x =
(y_i \land y_j) \land (y_i \land x)$.
The first half of this last expression is
bounded above by $z$, and the second half
is bounded above by $z'$, therefore
$y_i\land y'_j 
\leq z\land z' \leq y_j\land x = y'_j$.
Moreover, both $z$ and $z'$ being positive, and the target
being closed under intersection, 
ensures that $z\land z'$ is positive.

The induction basis case of appending a new $x$ to $N$
is handled in the same way:
the positive $z'$ obtained in the same manner
fulfills directly the condition 
$y_i \land x \leq z' \leq x$,
which is what we need.

Finally, the property that there exists a positive $z$ s.t.
$y_i \land y_j \leq z \leq y_j$ for every $i<j$ implies that each different $y_i, y_j$ must falsify a different implication of the target $T$. Suppose otherwise by way of contradiction that both counterexamples are falsifying the same implication $\alpha\rightarrow\beta$. Then, we would have that $\ones{\alpha}\leq y_i$ and $\ones{\alpha}\leq y_j$ so that $\ones{\alpha}\leq y_i\land y_j \leq z$. Since $z$ is positive, then $z\models \alpha\rightarrow\beta$ and so $\ones{\beta}\leq z\leq y_j$. Therefore $y_j\models\alpha\rightarrow\beta$ thus contradicting our assumption.
\end{proof}

To prove termination, it suffices to note that by the previous lemma, $N$ cannot be longer than the number of implications in the target. After each iteration, either an existing counterexample decreases in at least one bit (which can happen at most $n$ times for each existing counterexample), or a new one is added (which can happen at most $m$ times, where $m$ is the implication size of $T$).
Hence, the total number of equivalence queries issued is at most $nm + m +1 = O(nm)$. As to the number of closure queries, in each iteration we need to issue at most $m$ queries when checking intersections with existing members of $N$, which makes a total of $O(m^2n)$ closure queries. Notice that we could store and avoid the queries needed for building the hypothesis $hyp(N)$ and therefore we do not need to account for the extra $m$ queries (which in any case does not affect the asymptotic of the query count).
In terms of time, the outer loop is executed $O(mn)$ times, and each iteration has a cost of $mn$: the factor $m$ is due to looping over all $y_i\in N$, and the factor $n$ for the manipulations of vectors of length $n$, totaling a time complexity\footnote{This complexity depends on implementation details, but we assume these operations can be done in time linear with $n$, extra logarithmic factors could be hidden in a low-level implementation.} of $O(m^2 n^2)$.
\end{proof}

%

\begin{algorithm}
\begin{algorithmic}[1]
\STATE $N$ = [ ] \COMMENT{empty list}
\WHILE{$EQ(hyp(N)) = ($NO$,x)$}
\STATE\COMMENT{we will show below that $x$ is negative}
\FOR{$y_i\in N$, in order}
\STATE $y$ = $x \land y_i$
\IF{$y < y_i$ and $y < y^{\star}$}
\STATE $y_i$ = $y$
\PRINT
\ENDIF
\ENDFOR
\IF{no $y_i$ was changed}
\STATE add $x$ at the end of N
\ENDIF
\ENDWHILE
\end{algorithmic}
\caption{Learning from Closures Algorithm \H}
\label{fig:hstar}
\end{algorithm}

\subsection{The Horn Formula Obtained}

We prove now the main fact about algorithm \H,
characterizing its output. Most of the proof is discharged 
into the following technical lemma.

\begin{lemma}\label{lemma:leftsat}
At the time of issuing the equivalence query,
$hyp(N)$ is left-saturated.
\end{lemma}

\begin{proof}
For $hyp(N)$ to be left-saturated it is enough to show that $y_i \models \ones{y_j}\rightarrow \ones{y_j^{\star}}$ whenever $i\neq j$
since this implies that $y_i = y_i^{\bullet}$ or equivalently $y_i$ is closed with respect to $H\setminus H(\ones{y_i})$, where $H = hyp(N)$.

In order to show that an arbitrary $y_i\in N$ satisfies an arbitrary clause of $\ones{y_j}\rightarrow\ones{y_j^{\star}} \in hyp(N)$ whenever $i\neq j$, we proceed to show that $y_i \geq y_j$ implies $y_i \geq y_j^{\star}$ and so the implication is necessarily satisfied.

We assume, then, that $y_i \geq y_j$. If $i < j$, by Lemma~\ref{lemma:z} we know that 
$y_i \land y_j \leq z \leq y_j$, and so we have $y_j \leq z \leq y_j$ which is impossible since all $y_j$ are negative and $z$ is positive. Therefore, it must be the case that $i > j$ and then Lemma~\ref{lemma:z} guarantees that 
$y_j \land y_i = y_j \leq z \leq y_i$. Monotonicity of the closure operator implies that $y_j^{\star}\leq z^{\star}$ and so
$y_j^{\star}\leq z$ since $z^{\star} = z$. Finally, $y_j^{\star} \leq z \leq y_i$ implies $y_j^{\star}\leq y_i$ as required.
\end{proof}

\begin{theorem}
\label{t:hstargd}
The output of Algorithm \H\ is the GD basis of the target.
\end{theorem}

\begin{proof}
The output is the last hypothesis queried, which receives
a positive answer. By the previous lemma, all the antecedents
are left-saturated with respect to $hyp(N)$; but, as the 
answer is positive, $hyp(N)$ is equivalent to the target,
hence all the antecedents are left-saturated with respect
to the target. By construction, the right-hand sides of the
queries are always closures under the target. Hence, the
final query is a saturated definite Horn formula for the target.
As we have indicated earlier, there is a single saturated definite
Horn formula for any definite Horn theory: its GD basis. This is,
therefore, the output of the algorithm.
\end{proof}

\section{Relationships among Query Learning Models}


This section attempts to clarify the relationships between our algorithm 
and the previously published versions that work under slightly different learning 
models \cite{FrazierPitt,AngluinFrazierPitt,AriasBalcazarML}. 
The original AFP algorithm \cite{AngluinFrazierPitt,AriasBalcazarML} works under what we will refer to
as the \emph{Standard Query Model} which uses standard equivalence queries (SEQs) and standard membership queries (SMQs).

The algorithm LRN in \cite{FrazierPitt} works under the \emph{Entailment Query Model} which uses entailment membership queries (EMQs) and entailment equivalence queries (EEQs).
Entailment queries are somewhat more sophisticate versions
of the standard set-theoretic queries. In these queries, 
the role of assignments
is played here by clauses. In the entailment setting, a
membership query becomes a query to find out whether a
concrete clause provided by the learner is entailed by
the target. As  in
the set-theoretic setting, the equivalence query is  a Horn
formula, but the counterexample in case of nonequivalence
is a clause that is entailed by exactly one of the two
Horn formulas: the query and the target. This is, in fact,
the major difference with set-theoretic queries: the
entailment-based equivalence query does not return an 
$n$-bit vector but, instead, a clause.

Generally speaking,
there are two ways in which the relation between these algorithms
becomes apparent: 
the first one being that some queries can be directly simulated by others, and
so algorithms are the product of reductions;
but, also, there may be a way to specifically
run simulations of one particular algorithm within another,
even if the query protocol does not allow for direct simulation.

This section is divided into two parts. 
The first part (Section~\ref{sec:qsim}) will show direct simulations of several types of queries
by other query types. This type of reduction shows, in fact, the relationships among the three models considered (standard, entailment, and closure)
independent of the algorithm employed.
The second part (Section~\ref{sec:asim}) shows executions of the actual algorithms that lead to similar behaviors in the sense of having
identical evolution of intermediate hypotheses.

\subsection{Query Simulation}
\label{sec:qsim}

In this section we discuss cases where queries of one type
can be directly answered by (efficient) algorithms using another set of
queries. In this case, an algorithm working under one model can be directly made
to work under another model by using the appropriate query-answering algorithms as black boxes. 
These are, in fact, query model reductions.

In the following subsections we will detail several of these reductions. In some cases we will see how we can
simulate one type of query by its analogue under another model; in other cases, we may need both types of queries (membership and equivalence, for example)
to be able to simulate another query.
In our presentation, $T$ stands for the target Horn function.

\subsubsection{Entailment Queries Simulate Closure, Standard Membership, 
and Standard Equivalence Queries}

\paragraph{\bf EMQ $\implies$ CQ}
It is not hard to answer a CQ when EMQs are available.
Given $y$, to construct $y^{\star}$ (its closure with respect to $T$), we test, for each 
variable $b$ not in $\ones{y}$, whether $T\entails \ones{y}\implies b$ by means of EMQs. 
We include in $\ones{y^{\star}}$, apart from the variables that are already present in $\ones{y}$, all the $b$'s corresponding 
to positive answers from the EMQ. Clearly, this
constructs $y^{\star}$ with a linear cost in terms of EMQs.

\paragraph{\bf EMQ $\implies$ SMQ}
The same process provides for SMQs.
Indeed, a membership query on an assignment $x$ receives a
positive answer if and only if $x = x^{\star}$,
as per Proposition~\ref{p:closeds}. Essentially,
membership is negative if and only if 
there exists some variable $b$ not in $\ones{x}$ such that $T \entails \ones{x} \implies b$.
Again, the cost is linear.

\paragraph{\bf EMQ+EEQ $\implies$ SEQ}
We should note that
this case is just a detailed version of
Footnote 4 in  \cite{FrazierPitt}.
When answering an SEQ, unless the hypothesis is already equivalent to the target, we need
to return an assignment that satisfies the target but not the hypothesis or vice versa.
We first make an EEQ with the hypothesis and in return obtain a clause; from this
clause we need to find an assignment that distinguishes the target from our
hypothesis. We have two cases: it is a positive counterexample (entailed by the target but not by the hypothesis), or it is a negative counterexample (entailed by the hypothesis but not by the target).

The easier case is when the clause produced by the EEQ
is positive.
We transform it into a negative 
counterexample assignment~$x$ as follows.
Let $\alpha\implies b$ be the 
counterexample clause, so that $T$ entails $\alpha\implies b$
but $hyp(N)$ does not. There must be $x$ that satisfies $hyp(N)$ 
but does not satisfy $\alpha\implies b$, so that it cannot
satisfy $T$ because of the entailment from $T$. Such an $x$
is what we want.

How do we actually find it? To fail $\alpha\implies b$, 
it must satisfy $\alpha$, and also all the consequences of $\alpha$ 
under $hyp(N)$ in order to satisfy $hyp(N)$. The closure of $\bits{\alpha}$ 
under $hyp(N)$ (call it $w$) will do. Variable $b$ is not in that 
closure because the variables in the closure of $\alpha$ under 
$hyp(N)$ are exactly those variables $v$ for which $hyp(N)$ 
entails $\alpha\implies v$, and for $v=b$ it is not the case. 
Hence, $w$ fails $\alpha\implies b$, which is entailed by $T$, 
so $w$ cannot satisfy $T$, and satisfies $hyp(N)$ because it is 
a closure under it. So, in order to answer the EQ in this case one EEQ is enough
and the time complexity is what it takes to do forward-chaining with
the hypothesis, which can be done (when implemented carefully) in  linear time
in the number of implications in the hypothesis and the number of variables \cite{DowlingG84}.


The remaining case (counterexample clause entailed by the hypothesis but not by the target) 
can in fact be handled in the same way.
The only difference is that,
instead of closing $\bits{\alpha}$ under the hypothesis $hyp(N)$, we 
close it under the target, obtaining $\bits{\alpha}^{\star}$ via the
simulation of closures by EMQs.
So, in this case, one EEQ and a linear number of EMQs are needed in the 
worst case.

As a consequence of the ability of entailment queries to
implement both CQs and SEQs, from Theorems \ref{t:hstar} 
and \ref{t:hstargd} we obtain:

\begin{theorem}
\label{t:entailgd}
The following statements hold.
\begin{enumerate}
\item
\cite{FrazierPitt}
Horn theories are learnable from 
entailment queries in polynomial time.
\item
Further, such learning can be done so as to
output the GD basis of the target.
\end{enumerate}
\end{theorem}

\subsubsection{Closure and Equivalence Queries Simulate Entailment}

\paragraph{\bf CQ $\implies$ EMQ}
A CQ can easily simulate  a membership query of the 
entailment protocol. Given a clause $\alpha\implies v$, we can 
find out whether the target entails it by just asking for the
closure of the left-hand side and testing whether $v\in\alpha^{\star}$.
One single CQ suffices.

\paragraph{\bf CQ+SEQ $\implies$ EEQ}
For the simulation of an equivalence query of entailment, of course 
we resort to an SEQ; but we must transform
the assignment we get as counterexample into a counterexample clause for 
entailment. Given a negative counterexample assignment $x$, use a
CQ to obtain $x^{\star} \neq x$ and choose any variable $v$
that is true in $x^{\star}$ but not in $x$. Then, our counterexample
query is $\ones{x}\implies v$: as $x$ is positive for the query,
$v$ is not a consequence of $\ones{x}$ for the query, but it is
with respect to the target, as $v\in\ones{x^{\star}}$. Similarly, given
a positive counterexample $x$, that is, therefore, negative
for the query, we can find a counterexample clause $\ones{x}\implies v$ 
by finding some $v\notin\ones{x}$ that follows by forward chaining 
from $\ones{x}$ using the hypothesis in the query. Besides the SEQ, we spend at most one additional CQ
in this process. The total time would be $\mathcal{O}(nm)$ (here, $m$ is the implication size of the hypothesis). 

As a corollary, we obtain the following linear reductions among these three models:

\begin{corollary}
The following statements hold.
\begin{enumerate}
\item {\bf CQ+SEQ $\longleftrightarrow$ EMQ+EEQ.} The entailment and closure learning models are equivalent (up to a linear number of queries).
\item {\bf EMQ+EEQ $\implies$ SMQ+SEQ.} Entailment can simulate the standard protocol (up to a linear number of queries).
\item {\bf CQ+SEQ $\implies$ SMQ+SEQ.} The closure protocol can simulate the standard protocol (up to a linear number of queries).\end{enumerate}
\end{corollary}


It is worth noting that it is also possible to simulate closure queries (CQ) with the standard protocol (i.e., SEQ+SMQ $\implies$ CQ) by means of the following trivial (polynomial-query) reduction: when asked to compute a closure, we invoke the AFP algorithm of \cite{AngluinFrazierPitt} and once we discover the target
we can easily compute the closure. 
Notice that this takes $O(nm^2)$ queries so a polynomial reduction is indeed possible; however, we would like to see strictly better complexities. By transitivity, we would also obtain the (trivial) reduction SEQ+SMQ $\implies$ EEQ+EMQ using the same trick.
It remains an open question whether the reduction SEQ+SMQ $\implies$ CQ can be done with better query complexity. 

We can show, however, that having equivalence queries is necessary for the reduction to work.
That is, if equivalence queries are not available, then the reduction SMQ $\implies$ CQ  is not possible with a polynomial number of queries, as the following theorem shows:
\begin{theorem}\label{the:clos-hard}
Answering a CQ  may require an exponential number of SMQs.\end{theorem}
\begin{proof}
Let ${\cal F}$ be a family of Horn theories: ${\cal F} = \{f_x | x \in \B, x\neq 1^n \}$ where $f_x$  is the conjunction of two
parts: 
$$f_x = \bigwedge_{v \in \ones{x}}(\emptyset \implies v)  \land  \bigwedge_{w \notin \ones{x}}(w\implies \ones{1^n})$$
The first half of $f_x$ guarantees that any satisfying assignment $y$ is such that $x\leq y$, the second half guarantees that no assignment $y$ such that $x < y < 1^n$ satisfies $f_x$. Thus, each $f_x$ is satisfied by exactly two assignments: $x$ itself and the top $1^n$.

Now, we want to answer a CQ for the assignment $0^n$. For an arbitrary target $f_x\in {\cal F}$, the answer should be $x$. But obviously we do not know what
the target is and we need to answer the closure query by means of querying the standard membership query oracle. 
Answering the closure query correctly corresponds to identifying the target function $f_x$ among all candidates in $\cal F$ (of which there are $2^n-1$).
We use an adversarial strategy to show the exponential lower bound: all the answers to any membership query are going to be negative unless the input 
assignment to the query is $1^n$. Each query rules out only one potential target function and thus an exponential number of
queries is needed.
\end{proof}

In fact, Theorem~\ref{the:clos-hard} fits the general lower bounding scheme described in Lemma 2 of \cite{AngluinQaCL}. As a corollary we obtain that EMQs cannot be simulated with a polynomial number of SMQs either.

\begin{corollary}
Answering an EMQ  may require an exponential number of SMQs.
\end{corollary}

Table~\ref{tab:results} summarizes the results from this section. 

\begin{table}[h]
\begin{center}
\begin{tabular}{|l|c|c|}
\hline
Query simulation & Query complexity & Time complexity\\
\hline
SMQ $\implies$ EMQ & $\mathcal{O}(2^n)$ & $\mathcal{O}(2^n)$\\
SMQ $\implies$ CQ & $\mathcal{O}(2^n)$ & $\mathcal{O}(2^n)$\\
EMQ $\implies$ CQ & $\mathcal{O}(n)$ & $\mathcal{O}(n)$\\
EMQ $\implies$ SMQ & $\mathcal{O}(n)$ & $\mathcal{O}(n)$\\
CQ $\implies$ EMQ & 1 & $\mathcal{O}(n)$\\
CQ $\implies$ SMQ & 1 & $\mathcal{O}(n)$\\
SEQ+CQ $\implies$ EEQ & 1 SEQ + 1 CQ & $\mathcal{O}(nm)$\\
EEQ+EMQ $\implies$ SEQ & 1 EEQ + $\mathcal{O}(n)$ EMQ& $\mathcal{O}(nm)$\\
\hline
\end{tabular}
\end{center}
\caption{Relationship between different queries. As it is customary, $n$ stands for the number of propositional variables; $m$ stands for the implicational size of the input hypothesis.}
\label{tab:results}
\end{table}

Additionally, Table~\ref{tab:algs} summarizes the query and time complexities of the three algorithms that we compare in this paper for learning definite Horn theories. Note that while the worst-case complexities are the same for the three algorithms, there are cases where LRN and \H\ are going to be running faster than AFP, due to their slightly more powerful query models.

\begin{table}
\begin{center}
\begin{tabular}{|l|c|c|}
\hline
Algorithm & Query complexity & Time complexity\\
\hline
AFP & $\mathcal{O}(m^2n)$ SMQs \& $\mathcal{O}(mn)$ SEQs & $\mathcal{O}(m^2n^2)$\\
LRN & $\mathcal{O}(m^2n)$ EMQs \& $\mathcal{O}(mn)$ EEQs & $\mathcal{O}(m^2n^2)$\\
\H & $\mathcal{O}(m^2n)$ CQs \& $\mathcal{O}(mn)$ SEQs & $\mathcal{O}(m^2n^2)$\\
\hline
\end{tabular}
\end{center}
\caption{Query complexity and time complexity for AFP, LRN and \H.}
\label{tab:algs}
\end{table}

\subsection{Algorithm Run Simulation}
\label{sec:asim}

In this section we deal with the two remaining cases in which, as far as we know, the queries are not directly simulable (that is, not without learning the target first). We show that the full runs ``are'', in the sense that a run of one algorithm is 
embedded in some run of the other.

Namely, each of the algorithms that we consider here, even on the same
target, may exhibit different runs. More precisely, runs differ 
among them in which counterexamples are provided, and in which order.


\subsubsection{AFP Runs that Mimic \H\ Runs}

In the original membership and equivalence queries protocol, the AFP 
algorithm is not guaranteed to receive only negative counterexamples.
The reason is the lack of the closure query, that provides us with
positive examples.

In fact, each run of \H\ can be mimicked through a run of AFP
as follows. 
Fix the run of \H\ that receives the sequence of counterexamples
$x_1, x_2, \ldots, x_k$. We construct inductively a specific run 
of AFP that will receive this sequence of negative counterexamples
in the same order, plus positive ones as needed in between them. 
Consider the situation where it has just received the $j$-th
of them, with $j=0$ corresponding to the start of the algorithm.
The refinement process is the same in both cases, where the
tests for positive intersections are made via closures in one
algorithm and through direct memberships in the other. However,
at the point of constructing the query, one of the antecedents is
either new, or newly refined. For the rest, inductively, the closures
are the same as in the previous query, but AFP does not have available
the closure of the changed antecedent in order to use it as consequent.
It assumes the strongest possible consequent (or, in the variant in
\cite{AriasBalcazarML}, the strongest consequent compatible with positive
examples seen so far, so as to avoid the same counterexample to
show up over and over). 
In the specific run we are constructing, let $y_i$ be the
newly obtained antecedent. AFP might happen to hit upon
$y_i^{\star}$ on the basis of its available information:
then, it is asking exactly the next query of the \H\ run.
Otherwise, it is proposing too large a consequent for $y_i$: 
then, we give AFP the positive counterexample $\ones{y_i^{\star}}$, 
which fails the $y_i$ clause yet is positive, because
it is a closure. After this positive example, again the next
query is exactly \H's query. In either case, AFP proceeds 
and gets the $j+1$ negative counterexample.

Note that, along the way, a full formalization of this simulation
(which we consider unnecessary, as the intuition is clearly conveyed)
would provide an alternative proof of Theorem~\ref{t:hstargd},
as we get that every equivalence query (including its output)
made by \H\ is also a query made in some run of AFP on the
same target, and it is proved in \cite{AriasBalcazarML} that
all queries of AFP are saturated.

\paragraph{Query complexity} 
Looking closely at the closure queries made by \H, we note that they can be of two types: those that are issued 
to find out the ``right consequent'' of a newly added or updated implication in $hyp(N)$, and those that are issued
to check whether the intersections of negative counterexamples are positive or not (line 6 of Algorithm~\ref{fig:hstar}). 
If \H\ makes $a$ SEQs, then $a + b$ closure queries are made: $a$ accounting for the
former type (consequents of implications), and $b$ for the latter type. So \H\ issues $a$ SEQs an $a+b$ CQs.
Then, the corresponding run of AFP will make at most $2a$ SEQs (since we may need to feed all closures of 
antecedents as positive counterexamples in order to guide the algorithm towards the ``correct consequents'') and 
$b$ SMQs corresponding to checking the ``sign'' of interesections in order to know what antecedent to update.
So, both runs have the same total complexity, namely $2a + b$, although \H\ uses extra CQs to save on SEQs.

Of course, our discussion so far only applies to specific runs 
of AFP constructed
in that particular way. However, the properties of AFP proved
in \cite{AriasBalcazarML} show that all runs need to, eventually, 
identify the proper closures; in general, these necessary
positive counterexamples may not come at the place we are placing
them, but instead can come later; and, instead of the closures
$\ones{y_i^{\star}}$ that reduce the right-hand sides at once, we may reduce
them one bit at a time through several positive counterexamples. 
However, in a somewhat loose sense, we can say that AFP is implementing
the closure queries through potentially shuffled batches of positive 
counterexamples. This complicated procedure, whose most relevant
property is actually the goal of obtaining closures, makes us expect
that progress on the understanding of AFP, and, hopefully, either
a proof of optimality or an improvement of its query complexity,
could be obtained indirectly as a byproduct of the study of our
simpler, but essentially equivalent, algorithm \H.

\subsubsection{AFP Runs that Mimic Entailment Runs of LRN}
A similar development can be provided for mimicking runs
of the Learning from Entailment algorithm. We refrain from
getting into too much details here, as that would require, 
among other explanations, to review fully here the algorithm 
from \cite{FrazierPitt}. However, for the benefit of the
reader who knows, or plans to study soon, that algorithm,
we briefly point out how the simulation goes; it is quite
similar to the one in the previous subsection.

Fix the run that receives the sequence of counterexample
clauses $\alpha_1\to x_1$, $\alpha_2\to x_2$, \dots, $\alpha_k\to x_k$. 
We construct inductively a specific run of AFP that will receive 
a sequence of negative counterexamples, each corresponding, in
a precise sense, to each of these clauses. More precisely,
consider assignment $w_i$ defined as the closure of $\bits{\alpha_i}$
under the $i$-th hypothesis, to which $\alpha_i\to x_i$ itself
is a counterexample. Being a closure under the hypothesis,
$w_i$ is positive for it; however, it does not have 
variable $x_i$ set to 1, because the clause is not entailed
by the hypothesis, and this makes it a negative counterexample,
because the clause \emph{is} entailed by the target.

Again, each refinement process is identical in both algorithms,
so the difference is again upon constructing the new query.
The entailment algorithm can use entailment memberships to hit
the correct right-hand side. Instead, we consider a run of AFP
as before, where the appropriate positive counterexamples are
provided right away to lead the algorithm to the correct next
query of the simulated entailment run.

Again, we consider unnecessary to provide a full formalization 
of this simulation. However, since, again, all queries of AFP are 
saturated \cite{AriasBalcazarML}, we note that from such a full
formalization we can obtain a slightly stronger version of 
Theorem~\ref{t:entailgd}: in fact, the LRN algorithm already constructs the GD basis 
of the target, because every equivalence query there, including 
its output, is also a query made in some run of AFP on the
same target.

\section{Related Open Problems}

We would like to study how to extend our result to general Horn functions; the main difficulty being that it is not clear what closure means in the general case. In our previous work \cite{AriasBalcazarML} we come up with a pwm-reduction \cite{angluin1995won} that makes learning possible for general Horn under the model of standard equivalence and standard membership queries. It is left for future work to study a new reduction that works under the closure query model.

It also remains to prove or disprove the following relationships:
\begin{itemize}
	\item EMQ cannot be obtained with a linear number of SMQs and SEQs;
	\item CQ cannot be obtained with a linear number of SMQs and SEQs; 
	\item EEQ cannot be obtained with a polynomial number of SEQ;
	\item SEQ cannot be obtained with a polynomial number of EEQ.
\end{itemize}

\section*{Acknowledgments}

The authors are grateful to Montse Hermo for useful discussions.
We thank also the anonymous reviewers for their careful reading 
and thoughtful remarks, which have helped very much in improving 
the paper.

\bibliographystyle{elsarticle-num} 
\bibliography{bibfile}

\begin{thebibliography}{10}
\expandafter\ifx\csname url\endcsname\relax
  \def\url#1{\texttt{#1}}\fi
\expandafter\ifx\csname urlprefix\endcsname\relax\def\urlprefix{URL }\fi
\expandafter\ifx\csname href\endcsname\relax
  \def\href#1#2{#2} \def\path#1{#1}\fi

\bibitem{AngluinQaCL}
D.~Angluin, Queries and concept learning, Machine Learning 2~(4) (1987)
  319--342.

\bibitem{angluin1995won}
D.~Angluin, M.~Kharitonov, When won't membership queries help?, Journal of
  Computer and System Sciences 50~(2) (1995) 336--355.

\bibitem{BCGKL}
J.~L. Balc{\'a}zar, J.~Castro, D.~Guijarro, J.~K{\"o}bler, W.~Lindner, A
  general dimension for query learning, J. Comput. Syst. Sci. 73~(6) (2007)
  924--940.

\bibitem{AngluinFrazierPitt}
D.~Angluin, M.~Frazier, L.~Pitt, Learning conjunctions of {Horn} clauses,
  Machine Learning 9 (1992) 147--164.

\bibitem{GoldmanKwekScott}
S.~A. Goldman, S.~Kwek, S.~D. Scott, Learning from examples with unspecified
  attribute values, Inf. Comput. 180~(2) (2003) 82--100.

\bibitem{BDD}
S.~Ben-David, E.~Dichterman, Learning with restricted focus of attention, in:
  COLT, 1993, pp. 287--296.

\bibitem{FrazierPitt}
M.~Frazier, L.~Pitt, Learning from entailment: An application to propositional
  {Horn} sentences, in: ICML, Morgan Kaufmann, 1993, pp. 120--127.

\bibitem{GanterAE}
B.~Ganter, Attribute exploration with background knowledge, Theor. Comput. Sci.
  217~(2) (1999) 215--233.

\bibitem{BeDeTi06}
L.~Becerra-Bonache, A.~H. Dediu, C.~Tirn\u{a}uc\u{a}, Learning {DFA} from
  correction and equivalence queries, in: Y.~Sakakibara, S.~Kobayashi, K.~Sato,
  T.~Nishino, E.~Tomita (Eds.), ICGI, Vol. 4201 of Lecture Notes in Computer
  Science, Springer, 2006, pp. 281--292.

\bibitem{Tirn08}
C.~Tirn\u{a}uc\u{a}, A note on the relationship between different types of
  correction queries, in: A.~Clark, F.~Coste, L.~Miclet (Eds.), ICGI, Vol. 5278
  of Lecture Notes in Computer Science, Springer, 2008, pp. 213--223.

\bibitem{Tirn09}
C.~Tirn\u{a}uc\u{a}, Language learning with correction queries, Ph.D. thesis,
  Rovira i Virgili University, Tarragona, Spain (2009).

\bibitem{TirKob09}
C.~Tirn\u{a}uc\u{a}, S.~Kobayashi, Necessary and sufficient conditions for
  learning with correction queries, Theor. Comput. Sci. 410~(47-49) (2009)
  5145--5157.

\bibitem{AriasBalcazarALT}
M.~Arias, J.~L. Balc{\'a}zar, Canonical {Horn} representations and query
  learning, in: R.~Gavald{\`a}, G.~Lugosi, T.~Zeugmann, S.~Zilles (Eds.), ALT,
  Vol. 5809 of Lecture Notes in Computer Science, Springer, 2009, pp. 156--170.

\bibitem{AriasBalcazarML}
M.~Arias, J.~L. Balc{\'a}zar, Construction and learnability of canonical {Horn}
  formulas, Machine Learning 85~(3) (2011) 273--297.

\bibitem{Horn1956}
A.~Horn, On sentences which are true of direct unions of algebras, J. of
  Symbolic Logic 16 (1956) 14--21.

\bibitem{McKinsey1943}
J.~McKinsey, The decision problem for some classes of sentences without
  quantifiers, J. Symbolic Logic 8 (1943) 61--76.

\bibitem{KhardonRo1996}
R.~Khardon, D.~Roth, Reasoning with models, Artificial Intelligence 87~(1-2)
  (1996) 187 -- 213.

\bibitem{buning1999propositional}
H.~Kleine~B{\"u}ning, T.~Lettmann, {Propositional logic: deduction and
  algorithms}, Cambridge University Press, 1999.

\bibitem{GD}
J.~Guigues, V.~Duquenne, Familles minimales d'implications informatives
  resultants d'un tableau de donn\'ees binaires, Math. Sci. Hum. 95 (1986)
  5--18.

\bibitem{Maier1980}
D.~Maier, Minimum covers in relational database model, J. ACM 27 (1980)
  664--674.

\bibitem{Wild}
M.~Wild, A theory of finite closure spaces based on implications, Advances in
  Mathematics 108 (1994) 118--139.

\bibitem{DowlingG84}
W.~F. Dowling, J.~H. Gallier, Linear-time algorithms for testing the
  satisfiability of propositional {Horn} formulae, J. Log. Program. 1~(3)
  (1984) 267--284.

\end{thebibliography}

\end{document}